\documentclass[runningheads]{llncs}
\usepackage{graphicx}
\usepackage{amsmath}
\usepackage{amssymb}
\usepackage{array}
\usepackage{booktabs}
\usepackage{url}
\begin{document}
\title{Temporal Knowledge Graph Embedding Model based on Additive Time Series Decomposition}
%
%
\author{Chenjin Xu\inst{1}
\and
Mojtaba Nayyeri\inst{1}
\and
Fouad Alkhoury\inst{1}
\and
Hamed Yazdi\inst{1}\and
Jens Lehmann\inst{1,2}}
%
%
\institute{Smart Data Analytics Group, University of Bonn, Germany\\
\url{https://sda.tech/}\\
\email{\{xuc,nayyeri,shariat,jens.lehmann\}@cs.uni-bonn.de}
\and
 Enterprise Information Systems Department, Fraunhofer IAIS, Germany\\
 \email{\{jens.lehmann\}@iais.fraunhofer.de}
}
\maketitle              
\begin{abstract}
Knowledge Graph (KG) embedding has attracted more attention in recent years. Most KG embedding models learn from time-unaware triples. However, the inclusion of temporal information besides triples would further improve the performance of a KGE model. In this regard, we propose \textbf{ATiSE}\footnote{This paper has been accepted by ISWC2020}, a temporal KG embedding model which incorporates time information into entity/relation representations by using \textbf{A}dditive \textbf{Ti}me \textbf{Se}ries decomposition. Moreover, considering the temporal uncertainty during the evolution of entity/relation representations over time, we map the representations of temporal KGs into the space of multi-dimensional Gaussian distributions. The mean of each entity/relation embedding at a time step shows the current expected position, whereas its covariance (which is temporally stationary) represents its temporal uncertainty. Experimental results show that ATiSE significantly outperforms the state-of-the-art KGE models and the existing temporal KGE models on link prediction over four temporal KGs.

\keywords{	
Temporal knowledge graph  \and Knowledge representation and reasoning \and Time series decomposition.}
\end{abstract}
\section{Introduction}~\label{intro}
Knowledge Graphs (KGs) are being used for gathering and organizing scattered human knowledge into structured knowledge systems. 
YAGO~\cite{YAGO}, DBpedia~\cite{Dbpedia}, WordNet~\cite{WN} and Freebase~\cite{Freebase} are among existing KGs that have been successfully used in various applications including question answering, assistant systems, information retrieval, etc. 
In these KGs, knowledge can be represented as RDF triples (\textit{s, p ,o}) in which \textit{s} (subject) and \textit{o} (object) are entities (nodes), and \textit{p} (predicate) is the relation (edge) between them. 

KG embedding attempts to learn the representations of entities and relations in high-dimensional latent feature spaces while preserving certain properties of the original graph. Recently, KG embedding has become a very active research topic due to the wide ranges of downstream applications. Different KG embedding models have been proposed so far to efficiently learn the representations of KGs and perform KG completion as well as inferencing~\cite{TransE,KG2E,DISTMULT,ComplEx,RotatE,QuatE}. 

We notice that most of existing KG embedding models solely learn from time-unknown facts and ignore the useful temporal information in the KBs. In fact, there are many time-aware facts (or events) in some temporal KBs. For example, (\textit{Obama, wasBornIn, Hawaii}) happened at August 4, 1961. (\textit{Obama, presidentOf, USA}) was true from 2009 to 2017. These temporal KGs, e.g.\ Integrated Crisis Early Warning System (ICEWS)~\cite{ICEWS2015}, Global Database of Events, Language, and Tone (GDELT)~\cite{GDELT}, YAGO3~\cite{YAGO3} and Wikidata~\cite{Wikidata}, store such temporal information either explicitly or implicitly. 
Traditional KBE models such as TransE learn only from time-unknown facts. Therefore, they cannot distinguish entities with similar semantic meaning. 
For instance, they often confuse entities such as \textit{Barack Obama} and \textit{Bill Clinton} when predicting ${(?, \textit{presidentOf}, \textit{USA}, 2010)}$.

To tackle this problem, temporal KGE models~\cite{leblay,HyTE,TA-TransE} encode time information in their embeddings. TKGE models outperform traditional KGE models on link prediction over temporal KGs. 
It justifies that incorporation of time information can further improve the performance of a KGE model. Most existing TKGE models embed time information into a latent space, e.g.\, representing time as a vector. 
These models cannot capture some properties of time information such as the length of time interval as well as order of two time points. Moreover, these models ignore the uncertainty during the temporal evolution. We argue that the evolution of entity representations has randomness, because the features of an entity at a certain time are not completely determined by the past information. For example, (\textit{Steve Jobs, diedIn, California}) happened on 2011-10-05. The semantic characteristics of this entity should have a sudden change at this time point. However, due to the incompleteness of knowledge in KGs, this change can not be predicted only according to its past evolutionary trend. Therefore, the representation of \textit{Steve Jobs} is supposed to include some random components to handle this uncertainty, e.g.\, a Gaussian noise component.

In order to address the above problems, in this paper, we propose a temporal KG embedding model, ATiSE\footnote{The code is avaible at \url{https://github.com/soledad921/ATISE}}, which uses additive time series decomposition to capture the evolution process of KG representations. ATiSE fits the evolution process of an entity or relation as a multi-dimensional additive time series which composes of a trend component, a seasonal component and a random component. Our approach represents each entity and relation as a multi-dimensional Gaussian distribution at each time step to introduce a random component. The mean of an entity/relation representation at a certain time step indicates its current expected position, which is obtained from its initial representation, its linear change term, and its seasonality term. The covariance which describes the temporal uncertainty during its evolution, is denoted as a constant diagonal matrix for computing efficiency. Our contributions are as follows.

\begin{figure}
\vspace{-0.7cm}
\centering
\includegraphics[width=0.9\textwidth]{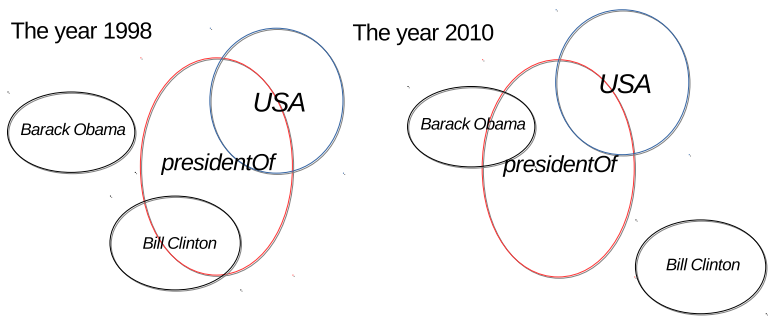}
\vspace{-0.4cm}
\caption{Illustration of the means and (diagonal) variances
of entities and relations in a temporal Gaussian Embedding Space. The labels indicate their position. In the representations, we might infer that \textit{Bill Clinton} was \textit{presidentOf} \textit{USA} in 1998 and \textit{Barack Obama} was \textit{presidentOf} \textit{USA} in 2010. } \label{instance.fig}
\vspace{-0.5cm}
\end{figure}

\begin{itemize}
    \item Learning the representations for temporal KGs is a relatively unexplored problem because most existing KG embedding models only learn from time-unknown facts. We propose ATiSE, a new KG embedding model to incorporate time information into the KG representations.
    \item We specially consider the temporal uncertainty during the evolution process of KG representations. Thus, we model each entity/relation as a Gaussian distribution at each time step. As shown in Figure~\ref{instance.fig}, the mean vectors of multi-dimensional Gaussian distributions of entities and relations indicate their position which changes over time and the covariance matrices indicate the corresponding temporal uncertainty. A symmetric KL-divergence between two Gaussian distributions is designed to compute the scores of facts for optimization.
    \item Different from the previous temporal KG embedding models which use time embedding to incorporate time information, ATiSE fits the evolution process of KG representations as a multi-dimensional additive time series. Our work establishes a previously unexplored connection between relational processes and time series analysis with a potential to open a new direction of research on reasoning over time.
    \item Our experimental results show that ATiSE significantly outperforms other TKG models and some state-of-the-art static KGE on link prediction over four TKG datasets.
\end{itemize}

The rest of the paper is organized as follows: In the section \ref{related work}, we first review related works; in the section \ref{method}, we introduce the architecture and the learning process of our proposed models; in the section \ref{experiments}, we compare the performance of our models with the state-of-the-art models; in the section \ref{conclusion}, we make a conclusion in the end of this paper.

\section{RELATED WORK}\label{related work}
A large amount of research has been done in KG embeddings~\cite{survey1}. A few examples of state-of-the-art KGE models include TransE~\cite{TransE}, TransH~\cite{TransH}, TransComplEx~\cite{TransComplEx}, RotatE~\cite{RotatE}, DistMult~\cite{DISTMULT}, ComplEx~\cite{ComplEx}, ComplEx-N3~\cite{Complex-N3} and QuatE~\cite{QuatE}. 

The above methods achieve good results on link prediction in KGs. However, these time-unaware KGE models have limitations on reasoning over TKGs. More concretely, given two quadruples with the same subjects, predicates, objects and different time stamps,  i.e., ${(\textit{Barack Obama}, \textit{presidentOf}, \textit{USA}, 2010)}$ and ${(\textit{Barack Obama}, \textit{presidentOf}, \textit{USA}, 2020)}$, static KGE models will model them with the same scores due to their ignorance of time information, while the validities of these two quadruples might be different.

Recent researches illustrate that the performances of KG embedding models can be further improved by incorporating time information in temporal KGs. 

TAE~\cite{TAE} captures the temporal ordering that exists between some relation types as well as additional common-sense constraints to generate more accurate link predictions.

TTransE~\cite{leblay} and HyTE~\cite{HyTE} adopt translational distance score functions and encode time information in the entity-relation low dimensional spaces with time embeddings and temporal hyperplanes. 

Know-Evolve~\cite{KnowEvolve2017} models the occurrence of a fact as a temporal point process. However, this method is built on
a problematic formulation when dealing with concurrent events, as shown in Section~\ref{baseline}.

TA-TransE and TA-DistMult\cite{TA-TransE} utilize recurrent neural networks to learn time-aware representations of relations and use standard scoring functions from TransE and DistMult. These models can model time information in the form of time points with or without some particular temporal modifiers, i.e., '\textit{occursSince}' and '\textit{occursUntil}'.

DE-SimplE~\cite{DE-SimplE} incorporates time information into diachronic entity embeddings and achieves the state of the art results on event-based TKGs. However, same as TA-TransE and TA-DistMult, DE-SimplE can not model facts involving time intervals shaped like $[2005, 2008]$.

Moreover, TEE~\cite{TEE} encodes representations of years into entity embeddings by aggregating the representations of the entities that occur in event-based descriptions of the years.

\section{OUR METHOD}\label{method}
In this section, we present a detailed description of our proposed method, ATiSE, which not only uses relational properties between entities in triples but also incorporates the associated temporal meta-data by using additive time series decomposition.
\subsection{Additive Time Series Embedding Model}\label{model}
\begin{figure}[h!]
\vspace{-0.5cm}
\centering
\includegraphics[width=0.6\textwidth]{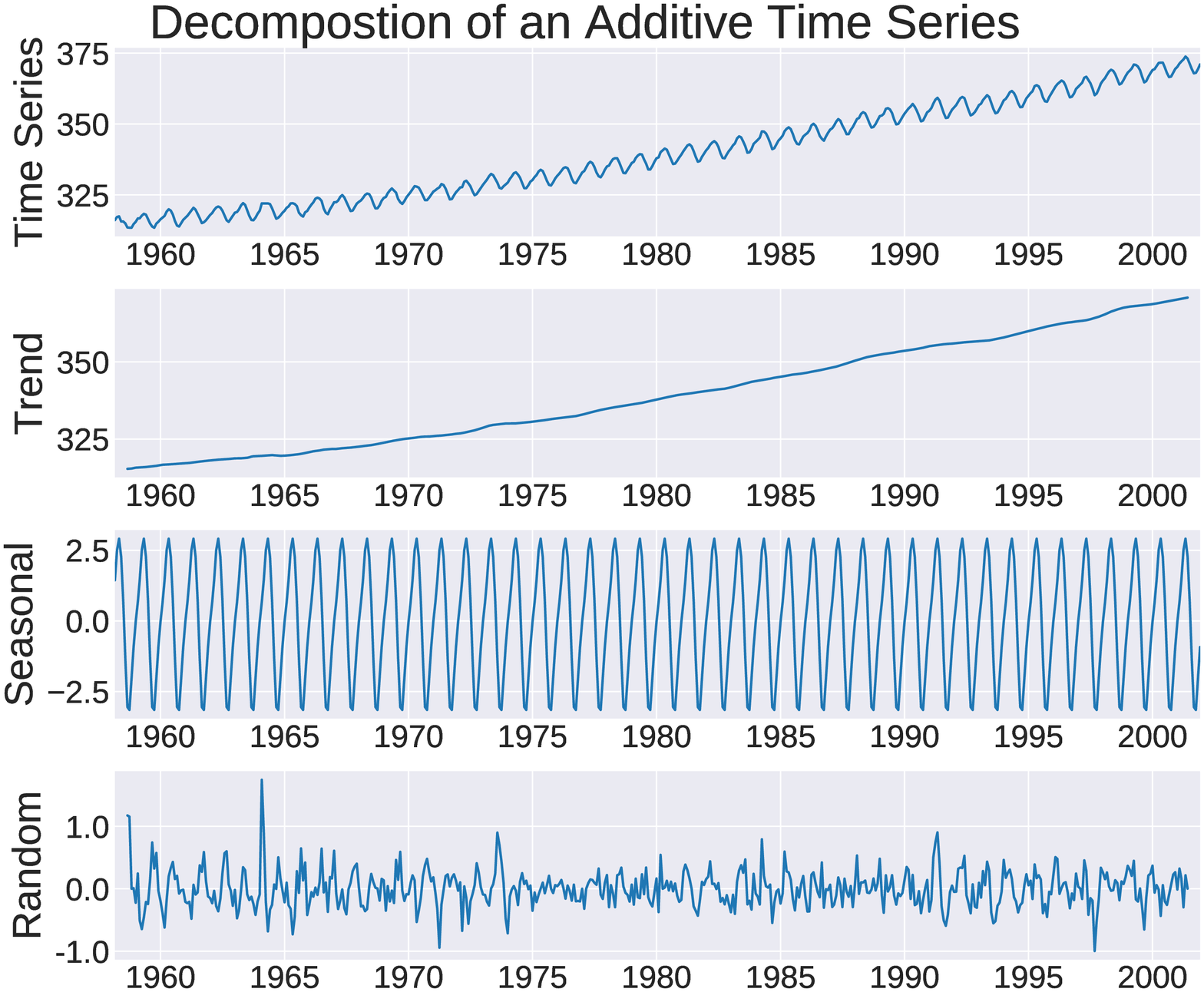} 
\caption{Illustration of additive time series decomposition.}
\label{fig:TimeSeries}
\vspace{-0.5cm}
\end{figure}

A time series is a series of time-oriented data. Time series analysis is widely used in many fields, ranging from economics and finance to managing production operations, to the analysis of political and social policy sessions~\cite{timeseries}. An important technique for time series analysis is additive time series decomposition. This technique decomposes a time series $Y_{t}$ into three components as follows,
\begin{equation}
    Y_{t}=T_{t}+S_{t}+R_{t}.
\end{equation}
where $T_{t}$, $S_{t}$ and $R_{t}$ denote the trend component, the seasonal component and the random component (i.e.\ “noise”), respectively. 
Figure~\ref{fig:TimeSeries} shows an instance of the additive time series decomposition of a time series. 

In our method, we regard the evolution of an entity/relation representation as an additive time series. For each entity/relation, we use a linear function and a Sine function to fit the trend component and the seasonal component respectively due to their simplicity. Considering the efficiency of model training, we model the irregular term by using a Gaussian noise instead of a moving average model (MA model)~\cite{ARIMA}, since training an MA model requires a global optimization algorithm which will lead to more computation consumption. 

To incorporate temporal information into traditional KGs, a new temporal dimension is added to fact triples, denoted as a quadruple $\left(s, p, o, t\right )$. It represents the creation of relationship edge \textit{p} between subject entity $s$, and object entity $o$ at time step \textit{t}. The score term $x_{spot}=f_{t}\left ( e_{s}, r_{p}, e_{o}\right )$ can represent the conditional probability or the confidence value of this event $x_{spot}$, where $e_{s}, e_{o} \in \mathbf{R}^{L_e}$, $r_{p} \in\mathbf{R}^{L_r} $ are representations of $s$, $o$ and $p$. In term of a long-term fact $\left(s, p, o, [t_{s}, t_{e}]\right )$, we consider it to be a positive triple for each time step between $t_s$ and $t_e$. $t_s$ and $t_e$ denote the start and end time during which the triple $\left(s, p, o\right )$ is valid.

At each time step, the time-specific representations of an entity $e_i$ or a relation $r_p$ should be updated as $e_{i,t}$ or $r_{p,t}$. Thus, the score of a quadruple $\left(s,p,o,t\right )$ can be represented as $x_{spot}=f_e\left ( e_{s,t}, r_{p,t}, e_{o,t} \right )$ or $x_{spot}=f_r\left ( e_{s}, r_{p,t}, e_{o} \right )$. We utilize additive time series decomposition to fit the evolution processes of each entity/relation representation as:
\begin{flalign}\label{eq1}
\begin{split}
    e_{i,t} = e_i+\alpha_{e,i} w_{e,i} t+\beta_{e,i}\text{sin}(2\pi\omega_{e,i} t)+\mathcal{N}(0,\Sigma_{e,i})\\
r_{p,t} = r_p+\alpha_{r,p} w_{r,p} t+\beta_{r,p}\text{sin}(2\pi\omega_{r,p} t)+\mathcal{N}(0,\Sigma_{r,p})\\
\end{split}
\end{flalign}
 where the $e_i$ and $r_p$ are the time-independent latent representations of the \textit{i}th entity which is subjected to $||e_i||_2=1 $ and the \textit{p}th relation which is subjected to $||r_p||_2=1 $. $e_{i}+\alpha_{e,i} w_{e,i} t$ and $r_p+\alpha_{r,p} w_{r,p} t$ are the trend components where the coefficients $|\alpha_{e,i}|$ and $|\alpha_{r,p}|$ denote the evolutionary rates of $e_{i,t}$ and $r_{p,t}$, the vectors $w_{e,i}$ and $w_{r,p}$ represents the corresponding evolutionary directions which are restricted to $||w_{e,i}||_2=||w_{r,p}||_2=1$. $\beta_{e,i}\text{sin}(2\pi\omega_{e,i} t)$ and $\beta_{r,p}\text{sin}(2\pi\omega_{r,p} t)$ are the corresponding seasonal components where $|\beta_{e,i}|$ and $|\beta_{r,p}|$ denote the amplitude vectors, $|\omega_{e,i}|$ and $|\omega_{r,p}|$ denote the frequency vectors. The Gaussian noise terms $\mathcal{N}(0,\Sigma_{e,i})$ and $\mathcal{N}(0,\Sigma_{r,p})$ are the random components, where $\Sigma_{e,i}$ and $\Sigma_{r,p}$ denote the corresponding diagonal covariance matrices.

In other words, for a fact $\left (s, p, o, t  \right )$, entity embeddings $e_{s,t}$ and $e_{o,t}$ obey Gaussian probability distributions: $\mathcal{P}_{s,t}\sim\mathcal{N}\left(\overline{e}_{s,t}, \Sigma_{s}\right)$ and $\mathcal{P}_{o,t}\sim\mathcal{N}\left(\overline{e}_{o,t}, \Sigma_{o}\right)$, where $\overline{e}_{s,t}$ and $\overline{e}_{o,t}$ are the mean vectors of $e_{s,t}$ and $e_{o,t}$, which do not include the random components. Similarly, the  predicate is represented as  $\mathcal{P}_{r,t}\sim\mathcal{N}\left(r_p, \Sigma_{r}\right)$.

Similar to translation-based KGE models, we consider the transformation result of ATiSE from the subject to the object to be akin to the predicate in a positive fact. We use the following formula to express this transformation: $\mathcal{P}_{s,t}-\mathcal{P}_{o,t}$, which corresponds to the probability distribution $\mathcal{P}_{e,t}\sim\mathcal{N}\left(\mu_{e,t}, \Sigma_{e}\right)$. Here, $\mu_{e,t}=\overline{e}_{s,t}-\overline{e}_{o,t}$ and $\Sigma_{e} = \Sigma_{s}+\Sigma_{o}$. Combined with the probability of relation $\mathcal{P}_{r,t}\sim\mathcal{N}\left(r_{p,t}, \Sigma_{r}\right)$, we measure the similarity between $\mathcal{P}_{e,t}$ and $\mathcal{P}_r$ to score the fact.

KL divergence is a straightforward method of measuring the similarity of two probability distributions. We optimize the following score function based on the KL divergence between the entity-transformed distribution and relation distribution~\cite{KLdivergence}.
\begin{align}\label{KLdivergence}
&x_{spot}=f_{t}\left ( e_{s},r_p,e_{o}\right )=
\mathcal{D_{KL}}(P_{r,t},P_{e,t})\nonumber\\
&=\int_{x\in \mathcal{R}^{{k_e}}} \mathcal{N}(x;r_{p,t},\Sigma_r)
\textup{log}\frac{\mathcal{N}(x;\mu_{e,t},\Sigma_e)}{\mathcal{N}(x;r_{p,t},\Sigma_r)}dx
\\
&=\frac{1}{2}\Big\{tr(\Sigma_{r}^{-1}\Sigma_e)+
(r_{p,t}-\mu_{e,t})^{T}\Sigma_r^{-1}(r_{p,t}-\mu_{e,t})\nonumber\\
&-\textup{log}\frac{det(\Sigma_{e})}{det(\Sigma_r)}-k_{e} \Big\}\nonumber
\end{align}
where, $tr(\Sigma)$ and $\Sigma^{-1}$ indicate the trace and inverse of the diagonal covariance matrix, respectively.

Since the computation of the determinants of the covariance matrices in Equation~\ref{KLdivergence} is time consuming, we define a symmetric similarity measure based on KL divergence to simplify the computation of the score function.
\begin{align}\label{symmetricKLdivergence}
f_{t}\left ( e_{s},r_p,e_{o}\right )=\frac{1}{2}(\mathcal{D_{KL}}(P_{r,t},P_{e,t})+\mathcal{D_{KL}}(P_{e,t},P_{r,t}))
\end{align}

 Considering the simplified diagonal covariance, we can compute the trace and inverse of the matrix simply and effectively for ATiSE. The gradient of log determinant is $\frac{\partial\textup{log} detA}{\partial A}=A^{-1}$, the gradient $\frac{\partial x^{T}A^{-1}y}{\partial A}=-A^{-T}xy^{T}A^{-T}$, and the gradient $\frac{\partial tr(X^{T}A^{-1}Y)}{\partial A}=-(A^{-1}YX^{T}A^{-1})^{T}$~\cite{petersen2008matrix}.

\subsection{Complexity}\label{complexity}

In Table~\ref{tb:complexity}, we summarize the scoring functions of several existing (T)KGE approaches and our models and compare their space complexities. $n_e$, $n_r$, $n_t$ and $n_{token}$ are numbers of entities, relations, time steps and temporal tokens used in~\cite{TA-TransE}; $d$ is the dimensionality of embeddings. $\langle x,y,z\rangle=\sum_{i}x_{i}y_{i}z_{i}$ denotes the tri-linear dot product; $\textsc{Re}(\cdot)$ denotes the real part of the complex embedding~\cite{ComplEx}; $\otimes$ denotes the Hamilton product between quaternion embeddings; ${\triangleleft}$ denotes the normalization of the quaternion embedding. $\mathcal{P}_t$ denotes the temporal projection for embeddings~\cite{HyTE}; $\textsc{LSTM}(\cdot)$ denotes an LSTM neural network; $[r_{p};t_{seq}]$ denotes the concatenation of the relation embedding and the sequence of temporal tokens~\cite{TA-TransE}; $\overrightarrow{e}$ and $\overleftarrow{e}$ denote the temporal part and untemporal part of a time-specific diachronic entity embedding $e^{t}$~\cite{DE-SimplE}; ${p}^{-1}$ denotes the inverse relation of ${p}$, i.e., $(s,p,o,t)\leftrightarrow (o,{p}^{-1},s,t)$. 

As shown in Table~\ref{complexity}, our model has the same space complexity and time complexity as static KGE models listed in Table~\ref{complexity} as well as DE-SimplE. On the other hand, the space complexities of TTransE, HyTE , TA-TransE or TA-DistMult will be higher than our models if $n_t$ or $n_{token}$ is much larger than $n_e$ and $n_r$.
\begin{table}
\vspace{-0.5cm}
\centering
\caption{Comparison of our models with several baseline models for space complexity.}
\resizebox{1\textwidth}{!}{
\begin{tabular}{|l|c|c|c|}
\hline
Model&Scoring Function&Space Complexity&Time Complexity\cr
\hline
TransE&$||e_s+r_p-e_o||$&$\mathcal{O}(n_{e}d+n_{r}d)$&$\mathcal{O}(d)$\cr
DistMult&$\langle e_{s},r_{p},e_{o}\rangle$&$\mathcal{O}(n_{e}d+n_{r}d)$&$\mathcal{O}(d)$\cr
ComplEx&$\textsc{Re}(\langle e_{s},r_{p},\overline{e}_{o}\rangle)$&$\mathcal{O}(n_{e}d+n_{r}d)$&$\mathcal{O}(d)$\cr
RotatE&$||e_s\circ r_p-e_o||$&$\mathcal{O}(n_{e}d+n_{r}d)$&$\mathcal{O}(d)$\cr
QuatE&$e_{s}\otimes r_{p}^{\triangleleft}\cdot e_{o}$&$\mathcal{O}(n_{e}d+n_{r}d)$&$\mathcal{O}(d)$\cr
\hline
TTransE&$||e_s+r_p+w_t-e_o||$&$\mathcal{O}(n_{e}d+n_{r}d+n_{t}d)$&$\mathcal{O}(d)$\cr
HyTE&$||P_{t}(e_{s})+P_{t}(r_{p})-P_{t}(e_{o})||$&$\mathcal{O}(n_{e}d+n_{r}d+n_{t}d)$&$\mathcal{O}(d)$\cr
TA-TransE&$||e_{s}+\textsc{LSTM}([r_{p};t_{seq}])-e_{o}||$&$\mathcal{O}(n_{e}d+n_{r}d+n_{token}d)$&$\mathcal{O}(d)$\cr
TA-DistMult&$\langle e_{s},\textsc{LSTM}([r_{p};t_{seq}]),e_{o}\rangle$&$\mathcal{O}(n_{e}d+n_{r}d+n_{token}d)$&$\mathcal{O}(d)$\cr
DE-SimplE&$\frac{1}{2}(\langle\overrightarrow{e}_{s}^{t},r_{p},\overleftarrow{e}_{o}^{t}\rangle+\langle\overrightarrow{e}_{0}^{t},r_{p^{-1}},\overleftarrow{e}_{s}^{t}\rangle)$&$\mathcal{O}(n_{e}d+n_{r}d)$&$\mathcal{O}(d)$\cr
\hline
ATiSE&$\mathcal{D_{KL}}(\mathcal{P}_{e,t},\mathcal{P}_{r,t})$&$\mathcal{O}(n_{e}d+n_{r}d)$&$\mathcal{O}(d)$\cr

\hline
\end{tabular}}
\vspace{-0.9cm}
\label{tb:complexity}
\end{table}

\subsection{Learning}\label{learn}
In this paper, we use the same loss function as the negative sampling loss proposed in~\cite{RotatE} for optimizing ATiSE. This loss function has been proved to be more effective than the margin rank loss function proposed in~\cite{TransE} on optimizing translation-based KGE models.
\begin{equation}~\label{loss}
\mathcal{L}=\sum_{t\in [T]}\sum_{\xi\in\mathcal{D}_{t}^{+}}\sum_{\xi^{'}\in\mathcal{D}_{t}^{-}}-\text{log}\ \sigma(\gamma-f_{t}(\xi))-\text{log}\ \sigma(f_{t}(\xi^{'})-\gamma)
\end{equation}%
where, $[T]$ is the set of time steps in the temporal KG, $\mathcal{D}_{t}^+$ is the set of positive triples with time stamp $t$, and $\mathcal{D}_{t}^-$ is the set of negative sample corresponding to $\mathcal{D}_{t}^+$. In this paper, we generate negative samples by randomly corrupting subjects or objects of the positives such as $(s^{'}, p, o, t)$ and $(s, p, o^{'}, t)$. Moreover, we adopt self-adversarial training proposed in~\cite{RotatE} and reciprocal learning used in~\cite{Complex-N3,DE-SimplE,QuatE} to further enhance the performances of our model.
To avoid overfitting, we add some regularizations while learning ATiSE. As described in the section \ref{model}, the norms of the original representations of entities and relations, as well as the norms of all evolutionary direction vectors, are restricted by 1. Besides, the following constraint is used for guaranteeing that the covariance matrices
are positive definite and of appropriate size when we minimize the loss:
\begin{flalign}\label{constraint}
\forall l \in \mathcal{E}\cup \mathcal{R},  c_{min}I\leq \Sigma_{l} \leq c_{max}I&
\end{flalign}
where, $\mathcal{E}$ and $\mathcal{R}$ are the set of entities and relations respectively, $c_{min}$ and $c_{max}$ are two positive constants. We use $\Sigma_{l}\leftarrow \max(c_{min},\min(c_{max},\Sigma_{l}))$ to achieve this regularization for diagonal covariance matrices. This constraint~\ref{constraint} for the covariance is considered during both the initialization and training process.
\begin{table}
\centering
\vspace{-0.3cm}
\begin{tabular}{ll}
\hline
  \multicolumn{2}{c}{\textbf{Algorithm:} The learning algorithm of \textbf{ATiSE}} \cr
  \hline
    &\textbf{input:} The training set $\mathcal{D}^{+}=\{(s,p,o,t)\}$, entity set $\mathcal{E},$ relation set $\mathcal{R}$, embedding\cr
   &dimensionality $d$, margin $\gamma$, batch size $b$, the ratio of negative samples over the  \cr
   &positives $\eta$, learning rate $lr$, restriction values $c_{min}$ and $c_{max}$ for covariance, and \cr
     & a score function $f_{t}(e_{s},r_{p},e_{o})$ where $s,o\in\mathcal{E}$, $p\in\mathcal{R}$.\cr 
     &\textbf{output:} Time-independent embeddings for each entity $e_i$ and relation $r_j$ (the\cr
    & mean vectors and the covariance matrices), the evolutionary rate $\alpha_{i}$and the\cr
    &evolutionary direction vector $w_{i}$ for each entity, where $i\in\mathcal{E}$, $j\in\mathcal{R}$.\cr
1.& \textbf{initialize} $e_{i},r_{j}\leftarrow$ uniform $(-\frac{6}{\sqrt{d}},\frac{6}{\sqrt{d}})$,  $i\in\mathcal{E}$, $j\in\mathcal{R}$\cr
  2.&$\qquad\qquad w_{e,i},w_{r,j}\leftarrow$ uniform $(-\frac{6}{\sqrt{d}},\frac{6}{\sqrt{d}})$,  $i\in\mathcal{E}$, $j\in\mathcal{R}$\cr
     3.&$\qquad\qquad \Sigma_{e,i}, \Sigma_{r,j}\leftarrow$ uniform $(c_{min},c_{max})$,  $i\in\mathcal{E}$, $j\in\mathcal{R}$\cr
     4.&$\qquad\qquad \omega_{e,i}, \omega_{r,j}\leftarrow$ uniform $(c_{min},c_{max})$,  $i\in\mathcal{E}$, $j\in\mathcal{R}$\cr
   5.&$\qquad\qquad \alpha_{e,i},\alpha_{r,j}\leftarrow$ uniform $(0,0)$,         $i\in\mathcal{E}$, $j\in\mathcal{R}$\cr
     6.&$\qquad\qquad \beta_{e,i},\beta{r,j}\leftarrow$ uniform $(0,0)$,         $i\in\mathcal{E}$, $j\in\mathcal{R}$\cr

7.&\textbf{loop}\cr
8.&$\quad e_{i}\leftarrow e_{i}/||e_{i}||_{2}$, $i\in\mathcal{E}$\cr
9.&$\quad r_{j}\leftarrow r_{j}/||r_{j}||_{2}$, $j\in\mathcal{R}$\cr
10.&$\quad w_{e,i}\leftarrow w_{e,i}/||w_{e,i}||_{2}$, $i\in\mathcal{E}$\cr
11.&$\quad w_{r,j}\leftarrow w_{r,j}/||w_{r,j}||_{2}$, $j\in\mathcal{R}$\cr
12.&$\quad \mathcal{D}^{+}_{b}\leftarrow$ sample($\mathcal{D}^{+}, b$) // sample a minibatch\cr
13.&$\quad$\textbf{for} $(s,p,o,t)\in\mathcal{D}^{+}_{\textit{b}}$  \textbf{do}\cr
14.&$\qquad \mathcal{D}_{b}^{-}=\{(s_{k}^{'},p,o_{k}^{'},t)\}_{k=1\dots\eta}$ // generate $\eta$ negative samples\cr
15.&$\quad$\textbf{end for}\cr
16.&$\quad$Update $e_i$, $w_i$, $\alpha_i$ and $r_j$ based on Equation~\ref{symmetricKLdivergence} and~\ref{loss} w.r.t.\cr
&$\quad \mathcal{L}=\sum_{\xi\in\mathcal{D}_{b}^{+}}\sum_{\xi^{'}\in\mathcal{D}_{b}^{-}}-\text{log}\ \sigma(\gamma-f_{t}(\xi))-\text{log}\ \sigma(f_{t}(\xi^{'})-\gamma)$\cr
17.&$\quad$regularize the covariances for each entity and relation based on Constraint~\ref{constraint},\cr
&$\quad \Sigma_{e,i}\leftarrow$$max(cmin,min(cmax,\Sigma_{e,i}))$, $i\in\mathcal{E}$\cr
&$\quad \Sigma_{r,j}\leftarrow$$max(cmin,min(cmax,\Sigma_{r,j}))$, $j\in\mathcal{R}$\cr
18.&\textbf{end loop}\cr
\hline
\end{tabular}
\vspace{-0.3cm}
\end{table}
\section{Experiment}~\label{experiments}
To show the capability of ATiSE, we compared it with some state-of-the-art KGE models and the existing TKGE models on link prediction over four TKG datasets. Particularly, we also did an ablation study to analyze the effect of the dimensionality of entity/relation embeddings and various components of the additive time series decomposition. 
\subsection{Datasets}\label{dataset}
As mentioned in section~\ref{intro}, common TKGs include ICEWS~\cite{ICEWS2015}, Wikidata~\cite{Wikidata} and YAGO3~\cite{YAGO3}. Four subsets of these TKGs are used as datasets in~\cite{TA-TransE}, i.e., ICEWS14, ICEWS05-15, YAGO15k and Wikidata11k. However, all of time intervals in YAGO15k and Wikidata11k only contain either start dates or end dates, shaped like \textit{'occursSince 2003'} or \textit{'occursUntil 2005'} while most of time intervals in Wikidata and YAGO are presented by both start dates and end dates. Thus, we prefer using YAGO11k and Wikidata12k released in~\cite{HyTE} instead of YAGO15k and Wikidta12k.
The statistics of the datasets used in this paper are listed in Table~\ref{tb:dataset}.
\begin{table}[ht]
\centering
\vspace{-0.5cm}
\caption{Statistics of datasets.
  }
  \resizebox{1\textwidth}{!}{ 
\begin{tabular}{|l|c|c|c|c|c|c|c|}
  \hline
   & \#Entities &\#Relations &\#Time Steps &Time Span& \#Training&\#Validation&\#Test\\
  \hline
  ICEWS14 & 6,869 & 230 &365& 2014&72,826&8,941&8,963 \\
  ICEWS05-15&10,094&251&4,017&2005-2015&368,962&46,275&46,092\\
  YAGO11k&10,623 &10 &70& -453-2844&16,408  &2,050 &2,051\\
  Wikidata12k&12,554 &24 &81&1709-2018&32,497&4,062&4,062\\
  \hline
\end{tabular}}
\vspace{-0.5cm}
  \label{tb:dataset}
\end{table}

\begin{table}[ht]
\centering
\vspace{-0.5cm}
\caption{Statistics of long-term facts
  }
  \resizebox{0.76\textwidth}{!}{ 
\begin{tabular}{|l|c|c|c|c|}
  \hline
    &\#Long-term Relations & \#Training&\#Validation&\#Test\\
  \hline
  YAGO11k&8 &12,579 &1,470&1,442\\
  Wikidata12k&20 &18,398 &2,194&2,200\\
  \hline
\end{tabular}}
  \label{tb:longterm facts}
  \vspace{-0.5cm}
\end{table}
ICEWS is a repository that contains political events with specific time annotations, e.g., (\textit{Barack Obama}, \textit{visits}, \textit{Ukraine}, \textit{2014-07-08}). ICEWS14 and ICEWS05-15 are subsets of ICEWS~\cite{ICEWS2015}, which correspond to the facts in 2014 and the facts between 2005 to 2015. These two datasets are filtered by only selecting the most frequently occurring entities in the graph~\cite{TA-TransE}. It is noteworthy that all of time annotations in ICEWS datasets are time points.

YAGO11k is a subset of YAGO3~\cite{YAGO3}. Different from ICEWS, a part of time annotations in YAGO3 are repsented as time intervals, e.g.\ (\textit{Paul Konchesky}, \textit{playsFor}, \textit{England national football team}, [2003-\#\#-\#\#, 2005-\#\#-\#\#]). Following the setting used in HyTE~\cite{HyTE}, we only deal with year level granularity by dropping the month and date information and treat timestamps as 70 different time steps in the consideration of the balance about numbers of triples in different time steps. For a time interval with the missing start date or end date, e.g., [2003-\#\#-\#\#, \#\#\#\#-\#\#-\#\#] representing 'since 2003', we use the first timestep or the last timestep to represent the missing start time or end time.

Wikidata12k is a subset of Wikidata~\cite{Wikidata}. Similar to YAGO11k, Wikidata12k contains some facts involving time intervals. We treat timestamps as 81 different time steps by using the same setting as YAGO11k.

As shown in Figure~\ref{tb:longterm facts}, most of facts in YAGO11k and Wikidata12k involve time intervals. For TKGE models, we discretized such facts $(s,p,o,[t_{s},t_{e}])$ involving multiple timesteps into multiple quadruples which only involve single timesteps, i.e., $\{(s,p,o,t_{s}),(s,p,o,t_{s+1}),\cdots,(s,p,o,t_{e})\}$, where $t_s$ and $t_e$ denote the start time and the end time.
\subsection{Evaluation Metrics}
We evaluate our model by testing the performances of our model on link prediction task over TKGs. This task is to complete a time-wise fact with a missing entity. For a test quadruple $(s, p, o, t)$, we generate corrupted triples by replacing $s$ or $o$ with all possible entities. We sort scores of all the quadruples including corrupted quadruples and the test quadruples and obtain the ranks of the test quadruples. For a test fact involving multiple time steps, e.g., $(s, p, o, [t_{s},t_{e}])$, the score of one corrupted fact $(s, p, o', [t_{s},t_{e}])$ is the sum of scores of multiple discreet quadruples, \{$(s,p,o',t_{s}),(s,p,o',t_{s+1}),\cdots,(s,p,o',t_{e})$\}. 

Two evaluation metrics are used here, i.e., Mean Reciprocal Rank and Hits@k. The Mean Reciprocal Rank (MRR) is the means of the reciprocal values of all computed ranks. And the fraction of test quadruples ranking in the top $k$ is called Hits@k. We adopt the time-wise filtered setting used in source code released by~\cite{DE-SimplE}. Different from the original filtered setting proposed in~\cite{TransE}, for a test fact $(s, p, o, t)$ or $(s, p, o, [t_{s},t_{e}])$, instead of removing all the triples that appear either in the training, validation or test set from the list of corrupted facts, we only filter the triples that occur at the time point $t$ or throughout the time interval $[t_{s},t_{e}]$ from the list of corrupted facts. This ensures that the facts that do not appear at $t$ or throughout $[t_{s},t_{e}]$ are still considered as corrupted triplets for evaluating the given test fact.
\subsection{Baselines}~\label{baseline}
We compare our approach with several state-of-the-art KGE approaches and existing TKGE approaches, including TransE~\cite{TransE}, DistMult~\cite{DISTMULT}, ComplEx-N3~\cite{Complex-N3}, RotatE~\cite{RotatE}, QuatE$^{2}$~\cite{QuatE}, TTransE~\cite{leblay}, TA-TransE, TA-DistMult~\cite{TA-TransE} and DE-SimplE~\cite{DE-SimplE}. ComplEx-N3 has been proven to have better performance than ComplEx~\cite{ComplEx} on FreeBase and WordNet datasets. And QuatE$^{2}$ has the best performances among all variants of QuatE as reported in~\cite{QuatE}.

As mentioned in Section~\ref{related work}, TA-TransE, TA-DistMult and DE-SimplE mainly focus on modeling temporal facts involving time points with or without some particular temporal modifiers, \textit{'occursSince'} and \textit{'occursUntil'}, and cannot model time intervals shaped like [2003-\#\#-\#\#, 2005-\#\#-\#\#]. Besides, DE-SimplE needs specific date information including year, month and day to score temporal facts, while most of time annotations in YAGO and Wikidataset only contain year-level information. Thus, we cannot test these three models on YAGO11k and Wikidataset15k.

We do not take Know-Evolve~\cite{KnowEvolve2017} as baseline model due to its problematic formulation and implementation issues. Know-Evolve uses the temporal point process to model the temporal evolution of each  entity. The intensity function of Know-Evolve (equation 3 in~\cite{KnowEvolve2017}) is defined as $\lambda_{r}^{s,o}(t|\overline{t})=f(g_{r}^{s,o}(\overline{t}))(t-\overline{t})$, where $g(\cdot)$ is a score function, $t$ is current time, and $\overline{t}$ is the most recent time point when either subject or object entity was involved in an event. This intensity function is used in inference to rank entity candidates. However,
they don’t consider concurrent event at the same time stamps, and thus $\overline{t}$ will become $t$ after one event. For example, we have events $event_{1} = (s, r, o_{1}, t_{1})$, $event_{2} = (s, r, o_{2}, t_{1})$. After $event_{1}$, $\overline{t}$ will become $t$
(subject $s$’s most recent time point), and thus the value of intensity function for $event_{2}$ will be 0. This
is problematic in inference since if $t$ = $\overline{t}$, then the intensity function will always be 0 regardless of
entity candidates. In their code, they give the highest ranks (first rank) for all entities including the ground truth object in this case, which we think is unfair since the scores of many entity candidates including the ground truth object might be 0 due to their formulation. It has been proven that the performances of Know-Evolve on ICEWS datasets drop down to almost zero after this issue fixed~\cite{RE-NET}.

\subsection{Experimental Setup}
 We used Adam optimizer to train our model and selected the optimal hyperparameters by early validation stopping according to MRR on the validation set. We restricted the maximum epoch to 5000. We fixed the mini-batch size $b$ as 512. We tuned the embedding dimensionalities $d$ in \{$100, 200, 300, 400, 500$\}, the ratio of negatives over positive training samples $\eta$ in \{$1, 3, 5, 10$\} and the learning rate $lr$ in \{$0.00003, 0.0001, 0.0003, 0.001$\}. The margins $\gamma$ were varied in the range \{1, 2, 3, 5, 10, 20, $\cdots$, 120\}. We selected the pair of restriction values $c_{min}$ and $c_{max}$ for covariance among \{(0.0001, 0.1), (0.003, 0.3), (0.005, 0.5), (0.01, 1)\}. The default configuration for ATiSE is as follows: $lr=0.00003$, $d=500$, $\eta=10$, $\gamma=1$, $(c_{min}, c_{max})=(0.005, 0.5)$. Below, we only list the non-default parameters: $\gamma=120$, $(c_{min}, c_{max})=(0.003, 0.3)$ on ICEWS14; $\gamma=100$, $(c_{min}, c_{max})=(0.003, 0.3)$ on ICEWS05-15.

\begin{table*}[h!]
\centering
    \caption{
    Link prediction results on ICEWS14 and ICEWS05-15. 
    *: results are taken from~\cite{TA-TransE}. $^{\diamond}$: results are taken from~\cite{DE-SimplE}. Dashes: results are unobtainable.
     The best results among all models are written bold.
    }
    \label{ICEWS results}

\begin{tabular}{|c|c|c|c|c|c|c|c|c|}
    \hline
  &\multicolumn{4}{c|}{ICEWS14}&\multicolumn{4}{c|}{ICEWS05-15}\cr 
  \hline
           Metrics&MRR&Hits@1&Hits@3&Hits@10&MRR&Hits@1&Hits@3&Hits@10\cr
\hline
 TransE*  &.280 &.094 &- &.637 &.294 &.090 &- &.663 \cr
        DistMult* &.439 &.323 &- &.672 &.456 &.337 &- &.691 \cr
        ComplEx-N3  &.467 &.347 &.527 &.716 &.481 &.362 &.535 &.729  \cr
        RotatE &.418 &.291 &.478 &.690 &.304 &.164 &.355 &.595 \cr
        QuatE$^{2}$ &.471 &.353 &.530 &.712 &.482 &.370 &.529 &.727 \cr
\hline
        TTransE$^{\diamond}$&.255 &.074 &- &.601 &.271 &.084 &- &.616 \cr
        HyTE$^{\diamond}$ &.297 &.108 &.416  &.655 &.316  &.116  &.445  &.681 \cr
        TA-TransE* &.275  &.095
        &-
        &.625 &.299 &.096 &- &.668  \cr
        TA-DistMult* &.477 &.363  &-  &.686  &.474 &.346  &- &.728   \cr
        DE-SimplE$^{\diamond}$ &.526  &.418 &.592  &.725 &.513  &\textbf{.392}  &.578  &.748 \cr
        \hline
        ATiSE &\textbf{.550} &\textbf{.436} &\textbf{.629} &\textbf{.750} &\textbf{.519} &.378 &\textbf{.606} &\textbf{.794}\cr
\hline
 
\end{tabular}
\end{table*}

\begin{table*}[h!]
\centering
    \caption{
    Link prediction results on YAGO11k and Wikidata12k. 
    The best results among all models are written bold.
    }
    \label{YAGO results}

\begin{tabular}{|c|c|c|c|c|c|c|c|c|}
    \hline
  &\multicolumn{4}{c|}{YAGO11k}&\multicolumn{4}{c|}{Wikidata12k}\cr 
       \hline   
           Metrics&MRR&Hits@1&Hits@3&Hits@10&MRR&Hits@1&Hits@3&Hits@10\cr
\hline
 TransE  &.100 &.015 &.138 &.244 &.178 &.100 &.192 &.339 \cr
        DistMult &.158 &.107 &.161 &.268 &.222 &.119 &.238 &.460 \cr
        ComplEx-N3  &.167 &.106 &.154 &.282 &.233 &.123 &.253 &.436  \cr
        RotatE &.167 &.103 &.167 &\textbf{.305} &.221 &.116 &.236 &.461 \cr
        QuatE$^{2}$ &.164 &.107 &.148 &.270 &.230 &.125 &.243 &.416 \cr
\hline
        TTransE&.108 &.020 &.150 &.251 &.172 &.096 &.184 &.329 \cr
        HyTE &.105  &.015 &.143  &.272 &.180  &.098 &.197  &.333 \cr
        \hline
        ATiSE &\textbf{.170} &\textbf{.110} &\textbf{.171} &.288 &\textbf{.280} &\textbf{.175} &\textbf{.317} &\textbf{.481}\cr
\hline
 
\end{tabular}

\end{table*}
\subsection{Experimental Results}
Table~\ref{ICEWS results} and~\ref{YAGO results} show the results for link prediction task. On ICEWS14 and ICEWS05-15, ATiSE outperformed all baseline models, considering MR, MRR, Hits@10 and Hits@1. Compared to DE-SimplE which is a very recent state-of-the-art TKGE model, ATiSE got improvement of 5\% on ICEWS14 regarding MRR, and improved Hits@10 by 3\% and 6\% on ICEWS14 and ICEWS05-15 respectively. On YAGO11k and Wikidata12k where time annotations in facts are time intervals, ATiSE surpassed baseline models regarding MRR, Hits@1, Hits@3. Regarding Hits@10, ATiSE achieved the state-of-the-art results on Wikidata12k and the second best results on YAGO11k. As mentioned in Section~\ref{baseline}, the results of TA-TransE, TA-DistMult and DE-SimplE on YAGO11k and Wikidata12k are unobtainable since they have difficulties in modeling facts involving time intervals in these two datasets.

A part of results listed on Table~\ref{ICEWS results} and~\ref{YAGO results} are obtained based on the implementations released in~\cite{RotatE,Complex-N3,HyTE}. We list the implementation details of some baseline models as follows:

\begin{itemize}
    \item We used the implementation released in~\cite{RotatE} to test RotatE on all four datasets, and DistMult on YAGO11k and Wikidata12k. The source code was revised to adopt the time-wise filtered setting. To search the optimal configurations for RotatE and DistMult, we followed the experimental setups reported in~\cite{RotatE} except setting the maximum dimensionality as 500 and the maximum negative sampling ratio as 10. The default optimal configuration for RotatE and DistMult is as follows: $lr=0.0001$, $b=1024$, $d=500$, $\eta=10$. Below, we only list the non-default parameters: for RotatE, the optimal margins are $\gamma=36$ on ICEWS14, $\gamma=48$ on ICEWS05-15, $\gamma=3$ on YAGO11k and $\gamma=6$ on Wikidata12k; for DistMult, the optimal regularizer weights are $r=0.00001$ on YAGO11k and Wikidata12k. 
      \item We used the implementation released in~\cite{Complex-N3} to test ComplEx-N3 and QuatE$^{2}$ on all four datasets. The source code was revised to adopt the time-wise filtered setting. To search the optimal configurations for ComplEx-N3 and QuatE$^{2}$, we followed the experimental setups reported in~\cite{Complex-N3} except setting the maximum dimensionality as 500. The default optimal configuration for ComplEx-N3 and QuatE$^{2}$ is as follows: $lr=0.1$, $d=500$, $b=1000$. Below, we list the optimal regularizer weights: for ComplEx-N3, $r=0.01$ on ICEWS14 and ICEWS05-15, $r=0.1$ on YAGO11k and Wikidata12k; for QuatE$^{2}$, $r=0.01$ on ICEWS14 and YAGO11k, $r=0.05$ on ICEWS05-15, $r=0.1$ on Wikidata.
      \item We used the implementation released in~\cite{HyTE} to test TransE, TTransE and HyTE on YAGO11k and Wikidata12k for obtaining their performances regarding MRR, Hits@1 and Hits@3. We followed the optimal configurations reported in~\cite{HyTE}. As shown in Table~\ref{YAGO results}, 
      Hits@10s of TransE and TTransE we got were better than those reported in~\cite{HyTE}.
\item  As shown in Table~\ref{ICEWS results}, other baseline results are taken from~\cite{TA-TransE,DE-SimplE}.
\end{itemize}
\subsection{Ablation Study}
In this work, we analyze the effects of the dimensionality and various components of entity/relation embeddings.
\begin{figure}
\vspace{-0.8cm}
\centering
\includegraphics[width=0.7\textwidth]{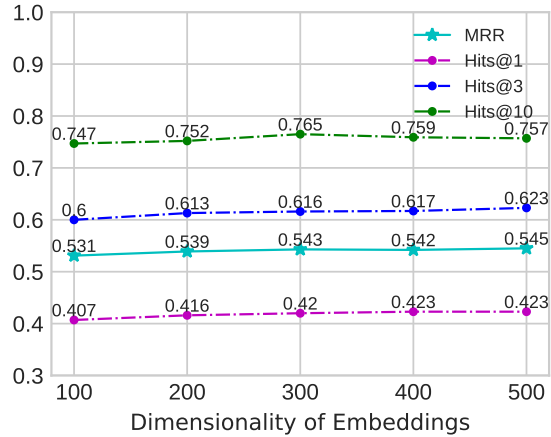}
\caption{Results for ATiSE with different embedding dimensionalities on ICEWS14.} \label{dim.fig}
\end{figure}

The embedding dimensionality is an important hyperparameter for each (T)KGE model. A high embedding dimensionality might be beneficial to boost the performance of a (T)KGE model. For instance, ComplEx-N3 and QuatE$^2$ achieved the state-of-the-art results on link prediction over static KGs with 2000-dimensional embeddings~\cite{Complex-N3,QuatE}. On the other hand, a lower embedding dimensionality will lead to less consumption on training time and memory space, which is quite important for the applications of (T)KGE models on large-scale datasets. Figure~\ref{dim.fig} shows the performances of ATiSE with different embedding dimensionalities on ICEWS14. With a same embedding dimensionality of 100 as DE-SimplE~\cite{DE-SimplE}, ATiSE still achieved the state-of-the-art results on ICEWS14. An ATiSE model with an embedding dimensionality of 100 trained on ICEWS14 had a memeory size of 14.2Mb while a DE-SimplE model and a QuatE$^2$ model with the same embedding dimensionlity had memory sizes of 13.3Mb and 12.4Mb. And the memory size of an ATiSE model increases linearly with its embedding dimensionality. Moreover, training an ATiSE model with an embedding dimensionality of 100 took 2.8 seconds per epoch on a single GeForce RTX2080, and an ATiSE with 500-dimensional embeddings took 3.7 seconds per epoch.

To analyze the effects of different components of entity/relation representation in ATiSE, we developed three comparison models, namely, ATiSE-SN, ATiSE-TN and ATiSE-TS, which exclude the trend component, seasonal component and the noise component respectively. The entity representations of these three comparison models are as follows:
\begin{flalign}\label{eq4}
\begin{split}
    e_{i,t}^{SN} &= e_i+\beta_{e,i}\text{sin}(2\pi\omega_{e,i} t)+\mathcal{N}(0,\Sigma_{e,i})\\
e_{i,t}^{TN} &= e_i+\alpha_{e,i} w_{e,i} t+\mathcal{N}(0,\Sigma_{e,i})\\
e_{i,t}^{TS} &=e_i+\alpha_{e,i} w_{e,i} t+\beta_{e,i}\text{sin}(2\pi\omega_{e,i} t)
\end{split}
\end{flalign}
For ATiSE-TS consisting of the trend component and the seasonal component, we used the translation-based scoring function~\cite{TransE} to measure the plausibility of the fact $\left (s, p, o, t  \right )$.
\begin{flalign}\label{eq5}
f_t^{TS}(e_s, r_p, e_o)=||e_{s,t}^{TS}+r_{p,t}^{TS}-e_{o,t}^{TS}||&
\end{flalign}
We report the MRRs and Hits@10 of ATiSE-SN, ATiSE-TN and ATiSE-TS on link prediction over ICEWS14 and YAGO11k. As shown in Table~\ref{tb3:AblationStudy}, we find that the removal of the trend component and the noise component had a remarkable negative effect on the performance of ATiSE on link prediction since the model could not address the temporal uncertainty of entity/relation representations without the noise component and the trend component contained the main time information. In ATiSE, different types of entities might have big difference in the trend component. For instance, we found that the embeddings of entities representing people, e.g., \textit{Barack Obama}, generally had higher evolution rates than those representing cities or nations, e.g., \textit{USA}.

\begin{table}[t]
\centering
    \caption{
    Link prediction results of ablation experiments. 
    }
\begin{tabular}{|l|c|c|c|c|c|c|c|c|}
  \hline
 Datasets&\multicolumn{4}{c|}{ICEWS14}&\multicolumn{4}{c|}{YAGO11K$_D$}\cr 
  \hline
            Metrics&MRR&Hits@1&Hits@3&Hits@10&MRR&Hits@1&Hits@3&Hits@10\cr
  \hline
        ATiSE-SN&.405 &.284 &.488 &.710 &.139 &.095 &.143 &.249   \cr
        ATiSE-TN&.536 &.407 &.626 &.771 &.167 &.105&.170&.282   \cr
        ATiSE-TS&.323 &.127 &.429 &.676 &.115 &.023 &.145 &.274  \cr
  \hline
        ATiSE&\textbf{.545} &\textbf{.423} &\textbf{.632} &\textbf{.757}&\textbf{.170} &\textbf{.110} &\textbf{.171} &\textbf{288}\cr
  \hline
\end{tabular}

    \label{tb3:AblationStudy}
\end{table}
ATiSE-TN performed worse than ATiSE on YAGO11k where facts involve time intervals. Different from ICEWS14 dataset which is an event-based dataset where all relations or predicates are instantaneous, there exist both short-term relations and long-term relations in YAGO11k. Adding seasonal components into evolving entity/relation representations is helpful to distinguish short-term patterns and long-term patterns in YAGO11k. It can be seen from Table~\ref{tb:relations} that short-term relations learned by ATiSE, e.g., \textit{wasBornIn}, generally had higher evolutionary rates, and their seasonal components had smaller amplitudes and higher frequencies than long-term relations, e.g., \textit{isMarriedTo}.
\begin{table}[ht]
\centering
\caption{Relations in YAGO11k and the mean step numbers of their duration time (TS), as well as the corresponding parameters learned from ATiSE, including the evolutionary rate $|\alpha_r|$, the mean amplitude $\overline{|\beta_r|}$ and the mean frequency $\overline{|\omega_r|}$ of the seasonal component for each relation.}\label{tb:relations}
\begin{tabular}{|l|c|c|c|c|}
\hline
  Relations&\#TS&$|\alpha_{r}|$&$\overline{|\beta_r|}$&$\overline{|\omega_r|}$ \cr
\hline
  \textit{wasBornIn}&1.0 &0.142 &0.000&1.032\cr
      \textit{worksAt}&18.7 &0.046 &0.058&0.294\cr
      \textit{playsFor}&4.7 &0.071 &0.046&0.766\cr
     \textit{hasWonPrize}&28.6 &0.010 &0.107&0.041\cr
  \textit{isMarriedTo}&16.5 &0.049 &0.076&0.090\cr
    \textit{owns}&24.9 &0.017 &0.088&0.101\cr
      \textit{graduatedFrom}&38.1 &0.016 &0.104&0.029\cr
        \textit{deadIn}&1.0 &0.249 &0.006&0.897\cr
        \textit{isAffliatedTo}&25.8 &0.014 &0.049&0.126\cr
        \textit{created}&27.1 &0.011 &0.040&0.087\cr
\hline
\end{tabular}
\end{table}

\section{CONCLUSION}\label{conclusion}
We introduce ATiSE, a temporal KGE model that incorporates time information into KG representations by fitting the temporal evolution of entity/relation representations over time as additive time series. Considering the uncertainty during the temporal evolution of KG representations, ATiSE maps the representations of temporal KGs into the space of multi-dimensional Gaussian distributions. The covariance of an entity/relation representation represents its randomness component. Experimental results demonstrate that our method significantly outperforms the state-of-the-art methods on link prediction over four TKG benchmarks.

Our work establishes a previously unexplored connection between relational processes and time series analysis with a potential to open a new direction of research on reasoning over time. In the future, we will explore to use more sophisticated models to model different components of relation/entity representations, e.g., an ARIMA model for the noise component and a polynomial model for the trend component.

\smallskip \noindent
{\footnotesize \textbf{Acknowledgements}. 
This work is supported by the CLEOPATRA project (GA no.~812997), the German national funded BmBF project MLwin and the BOOST project.}
\vspace*{-1em}
\bibliographystyle{splncs04}
\bibliography{iswc}

\begin{thebibliography}{10}
\providecommand{\url}[1]{\texttt{#1}}
\providecommand{\urlprefix}{URL }
\providecommand{\doi}[1]{https://doi.org/#1}

\bibitem{Dbpedia}
Auer, S., Bizer, C., Kobilarov, G., Lehmann, J., Cyganiak, R., Ives, Z.:
  Dbpedia: A nucleus for a web of open data. The semantic web pp. 722--735
  (2007)

\bibitem{TEE}
Bianchi, F., Palmonari, M., Nozza, D.: Towards encoding time in text-based
  entity embeddings. In: International Semantic Web Conference. pp. 56--71.
  Springer (2018)

\bibitem{Freebase}
Bollacker, K., Evans, C., Paritosh, P., Sturge, T., Taylor, J.: Freebase: a
  collaboratively created graph database for structuring human knowledge. In:
  Proceedings of the 2008 ACM SIGMOD international conference on Management of
  data. pp. 1247--1250. AcM (2008)

\bibitem{TransE}
Bordes, A., Usunier, N., Garcia-Duran, A., Weston, J., Yakhnenko, O.:
  Translating embeddings for modeling multi-relational data. In: Advances in
  Neural Information Processing Systems. pp. 2787--2795 (2013)

\bibitem{HyTE}
Dasgupta, S.S., Ray, S.N., Talukdar, P.: Hyte: Hyperplane-based temporally
  aware knowledge graph embedding. In: Proceedings of the 2018 Conference on
  Empirical Methods in Natural Language Processing. pp. 2001--2011 (2018)

\bibitem{Wikidata}
Erxleben, F., G{\"u}nther, M., Kr{\"o}tzsch, M., Mendez, J.,
  Vrande{\v{c}}i{\'c}, D.: Introducing wikidata to the linked data web. In:
  International Semantic Web Conference. pp. 50--65. Springer (2014)

\bibitem{TA-TransE}
Garc{\'\i}a-Dur{\'a}n, A., Duman{\v{c}}i{\'c}, S., Niepert, M.: Learning
  sequence encoders for temporal knowledge graph completion. In: EMNLP (2018)

\bibitem{DE-SimplE}
Goel, R., Kazemi, S.M., Brubaker, M., Poupart, P.: Diachronic embedding for
  temporal knowledge graph completion. In: AAAI (2020)

\bibitem{KG2E}
He, S., Liu, K., Ji, G., Zhao, J.: Learning to represent knowledge graphs with
  gaussian embedding. In: Proceedings of the 24th ACM International on
  Conference on Information and Knowledge Management. pp. 623--632. ACM (2015)

\bibitem{ARIMA}
Ho, S., Xie, M.: The use of arima models for reliability forecasting and
  analysis. Computers \& industrial engineering  \textbf{35}(1-2),  213--216
  (1998)

\bibitem{TAE}
Jiang, T., Liu, T., Ge, T., Sha, L., Chang, B., Li, S., Sui, Z.: Towards
  time-aware knowledge graph completion. In: Proceedings of COLING 2016, the
  26th International Conference on Computational Linguistics: Technical Papers.
  pp. 1715--1724 (2016)

\bibitem{RE-NET}
Jin, W., Jiang, H., Qu, M., Chen, T., Zhang, C., Szekely, P., Ren, X.:
  Recurrent event network: Global structure inference over temporal knowledge
  graph. arXiv: 1904.05530  (2019)

\bibitem{Complex-N3}
Lacroix, T., Usunier, N., Obozinski, G.: Canonical tensor decomposition for
  knowledge base completion. In: International Conference on Machine Learning.
  pp. 2869--2878 (2018)

\bibitem{ICEWS2015}
Lautenschlager, J., Shellman, S., Ward, M.: Icews event aggregations (2015).
  \doi{10.7910/DVN/28117}, \url{https://doi.org/10.7910/DVN/28117}

\bibitem{leblay}
Leblay, J., Chekol, M.W.: Deriving validity time in knowledge graph. In:
  Companion of the The Web Conference 2018 on The Web Conference 2018. pp.
  1771--1776. International World Wide Web Conferences Steering Committee
  (2018)

\bibitem{GDELT}
Leetaru, K., Schrodt, P.A.: Gdelt: Global data on events, location, and tone,
  1979--2012. In: ISA annual convention. vol.~2, pp. 1--49. Citeseer (2013)

\bibitem{YAGO3}
Mahdisoltani, F., Biega, J., Suchanek, F.M.: Yago3: A knowledge base from
  multilingual wikipedias. In: CIDR (2013)

\bibitem{WN}
Miller, G.A.: WordNet: An electronic lexical database. MIT press (1998)

\bibitem{timeseries}
Montgomery, D.C., Jennings, C.L., Kulahci, M.: Introduction to time series
  analysis and forecasting. John Wiley \& Sons (2015)

\bibitem{TransComplEx}
Nayyeri, M., Xu, C., Yaghoobzadeh, Y., Yazdi, H.S., Lehmann, J.: Toward
  understanding the effect of loss function on the performance of knowledge
  graph embedding  (2019)

\bibitem{petersen2008matrix}
Petersen, K.B., Pedersen, M.S., et~al.: The matrix cookbook. Technical
  University of Denmark  \textbf{7}(15), ~510 (2008)

\bibitem{YAGO}
Suchanek, F.M., Kasneci, G., Weikum, G.: Yago: a core of semantic knowledge.
  In: Proceedings of the 16th international conference on World Wide Web. pp.
  697--706. ACM (2007)

\bibitem{RotatE}
Sun, Z., Deng, Z.H., Nie, J.Y., Tang, J.: Rotate: Knowledge graph embedding by
  relational rotation in complex space. In: ICLR (2019)

\bibitem{KnowEvolve2017}
Trivedi, R., Dai, H., Wang, Y., Song, L.: Know-evolve: Deep temporal reasoning
  for dynamic knowledge graphs. In: ICML (2017)

\bibitem{ComplEx}
Trouillon, T., Welbl, J., Riedel, S., Gaussier, {\'E}., Bouchard, G.: Complex
  embeddings for simple link prediction. In: Proceedings of ICML (2016)

\bibitem{survey1}
Wang, Q., Mao, Z., Wang, B., Guo, L.: Knowledge graph embedding: A survey of
  approaches and applications. IEEE Transactions on Knowledge and Data
  Engineering  \textbf{29}(12),  2724--2743 (2017)

\bibitem{TransH}
Wang, Z., Zhang, J., Feng, J., Chen, Z.: Knowledge graph embedding by
  translating on hyperplanes. In: AAAI. pp. 1112--1119. Citeseer (2014)

\bibitem{DISTMULT}
Yang, B., Yih, W.t., He, X., Gao, J., Deng, L.: Embedding entities and
  relations for learning and inference in knowledge bases. In: ICLR. p.~12
  (2015)

\bibitem{KLdivergence}
Yu, D., Yao, K., Su, H., Li, G., Seide, F.: Kl-divergence regularized deep
  neural network adaptation for improved large vocabulary speech recognition.
  In: 2013 IEEE International Conference on Acoustics, Speech and Signal
  Processing. pp. 7893--7897. IEEE (2013)

\bibitem{QuatE}
Zhang, S., Tay, Y., Yao, L., Liu, Q.: Quaternion knowledge graph embeddings.
  In: Advances in Neural Information Processing Systems. pp. 2731--2741 (2019)

\end{thebibliography}

\end{document}